# A Conjecture on a Fundamental Trade-Off between Certainty and Scope in Symbolic and Generative AI

Luciano Floridi[1,2]
[1] Digital Ethics Center, Yale University, 85 Trumbull St, New Haven, CT 06511
[2] Department of Legal Studies, University of Bologna, Via Zamboni 27/29, 40126 Bologna, Italy

## 1. Introduction

Since the mid-twentieth century, two broad paradigms have shaped AI research. Classical symbolic systems, grounded in mathematical logic and rule manipulation, can deliver proofs of the correctness of their outcomes (henceforth also referred to as *certainty*) on narrowly constrained, fully formalised inputs (Russell and Norvig 2022). By contrast, modern generative systems—most notably large transformer-based architectures—perform high-dimensional mappings from data to rich, informational outputs (henceforth also referred to as *scope*), but provide only statistical performance guarantees about their outcomes (Goodfellow, Bengio, and Courville 2016; Bubeck *et al.* 2023). Figure 1 gives a brief overview of the key differences between symbolic and generative AI.

| Dimension | Symbolic AI (≈ 1956–2010) | Generative AI (≈ 2017–present) |
|---|---|---|
| Knowledge Source | Manually engineered rules and ontologies; closed-world assumption. | Self-supervised gradient learning from petabyte-scale corpora; open-world assumption. |
| Internal Representation | Discrete, logic-based symbols with explicit semantics. | Continuous, high-dimensional embeddings with distributed semantics. |
| Inference Mechanism | Deterministic search, theorem proving, or planning over explicit rules; brittle under noise. | Probabilistic sampling from learned distributions; robust to noise, capable of few-shot generalisation. |
| Input Modality | Requires pre-symbolised, fully structured data. | Inputs unstructured data (e.g., text, images, audio, code). |
| Output Modality | Yields symbolic proofs, plans, or database updates; struggles with high-entropy artefacts. | Generates coherent high-entropy artefacts (e.g., paragraphs, images, executables). |
| Typical Strengths | Formal verification, game-tree search, expert-system diagnostics in narrow domains. | Multimodal perception-generation, open-domain dialogue, style transfer, creative synthesis. |
| Typical Limitations | Poor perceptual grounding; no automatic feature learning; limited scalability. | Opaque internal reasoning; controllability and factual fidelity remain open problems. |

*Figure 1 Scope Differential: Classical Symbolic AI vs Modern Generative AI*

It has long been suspected that these different affordances are mutually limiting: the broader the coverage of unstructured reality, the less room there is for deductively watertight guarantees, that is, certainty about the error-free nature of the outcomes, and *vice versa*. This article formalises this intuition as a quantitative conjecture. If proved, the conjecture would explain why provably "zero-error" symbolic mechanisms cannot simultaneously serve as open-world multimodal generators,[1] and why large-scale generative models must admit an irreducible probability of error or hallucination. Conversely, if disproved—logico-mathematically or empirically (i.e., by a concrete counter-example)—the failure of the conjecture would expose an as-yet unexplored design space where provable certainty (about the reliability of the outcomes) and broad data scope coexist. As of now, however, the conjecture remains open. Whether it will eventually yield to a rigorous proof, be refuted, or simply persist as an empirical regularity is an open and consequential question.[2]

## 2. Formalising the conjecture

As we shall see in section 3, several results suggest that no AI system can simultaneously guarantee flawless correctness of its output on every possible input *and* operate over the full richness of unstructured, real-world data. A system that promises demonstrable certainty must confine itself to a well-defined, low-entropy niche, much as a pocket calculator is infallible only for bounded-precision arithmetic. By contrast, a system that can accept photographs, prose, or video as input and produce poems, protein folds, or executable code as output seems forced to abandon the possibility of provably perfect performance. Informally stated, the conjecture posits a strict trade-off: the closer one draws to absolute certainty (error probability → 0), the more one must contract the scope of admissible data and increasingly powerful applications; conversely, widening that scope inevitably reintroduces some irreducible uncertainty or imperfection about the output. Formally, one can state the conjecture as follows.[3]

Let an AI mechanism $M$ compute a total function $f_M : I \rightarrow O$ from inputs to outputs. Define two scalar measures:

**Epistemic certainty:** $C(M) := 1 - \sup_{x \in I} Pr\,[f_M(x) \neq Spec\,(x)]$
where $Spec\,(x)$ denotes the ground-truth specification for input $x$, and the supremum reflects the worst-case error probability over the input space. More informally, $\boldsymbol{Spec}(\boldsymbol{x})$ is the correct answer that the AI should give for any input $\boldsymbol{x}$, and the "supremum" part means one is looking at the highest chance of making an error across all possible inputs, basically finding where the AI is most likely to get things wrong. Thus, $C(M)$ quantifies the tightest provable upper bound on the mechanism's maximum error rate. A value of $C(M) = 1$ entails a formal guarantee that $f_M(x) = \ Spec\ (x)$ for all $x$.

[1] Open-world multimodal generators are AI systems designed to process and produce content across multiple modalities (such as text, images, audio, and video) while operating in environments that are unbounded, diverse, and potentially contain novel scenarios not explicitly covered during training.

[2] I am glad I do not have to commit to one of the alternatives in this article, but if forced, I suspect I would opt for the "empirical regularity" alternative.

[3] A previous version of this paper contained errors in the formalization. I apologise. I am deeply grateful to Alberto Messina for bringing these issues to my attention and enabling me to correct them.

**Mapping scope:** $\boldsymbol{S(M) := K(I) + K(O)}$
is the joint Kolmogorov complexity of the input and output spaces, serving as a proxy for the richness or information-theoretic breadth of the domain.

To illustrate this concept intuitively, consider two contrasting examples. A simple symbolic calculator, performing arithmetic operations on single-digit integers, has a highly constrained input-output domain; thus, both $\boldsymbol{K}(\boldsymbol{I})$ and $\boldsymbol{K}(\boldsymbol{O})$ are extremely small, resulting in a low value of $\boldsymbol{S}(\boldsymbol{M})$. Conversely, a large-scale generative language model, such as GPT-4, accepts, for example, a vast range of natural-language prompts and produces diverse, complex textual outputs, leading to very large $\boldsymbol{K}(\boldsymbol{I})$ and $\boldsymbol{K}(\boldsymbol{O})$, and consequently, a high value of $\boldsymbol{S}(\boldsymbol{M})$. This intuitive contrast underscores how Kolmogorov complexity naturally aligns with our intuitive sense of domain breadth or informational richness.[4]

**Conjecture**
There exists a universal constant $k > 0$, independent of any specific system, such that every sufficiently expressive[5] AI mechanism $M$ satisfies the following fundamental inequality: $C(M) \cdot S(M) \leq k$.
Intuitively, this means that, as certainty about AI systems' outcomes increases, scope must shrink. As AI systems' scope expands to encompass complex, unstructured domains and their corresponding powerful applications, achieving perfect certainty about the reliability of their results becomes unattainable. The conjecture is that no system can escape this trade-off. Figure 2 provides a visual representation.

[4] Interestingly, Alberto Messina has suggested that using Shannon's information instead of Kolmogorov's complexity is computationally and pragrmatically preferable. I agree, but it is not my idea and it would not be fair to adopt it here. I look forward to reading his forthcoming contribution.
[5] Here, the phrase "sufficiently expressive" clarifies that the conjecture targets AI systems of practical and theoretical interest rather than trivial computational devices with extremely restricted domains. The constant *k* is a universal, positive constant strictly less than 1, independent of any particular AI system. Its exact magnitude is a crucial theoretical and empirical question. A larger *k* would mean the product can approach unity more closely, implying a less pronounced trade-off, whereas a smaller *k* yields a more severe trade-off. Future theoretical or empirical work will clarify the precise range and realistic magnitude of *k*.

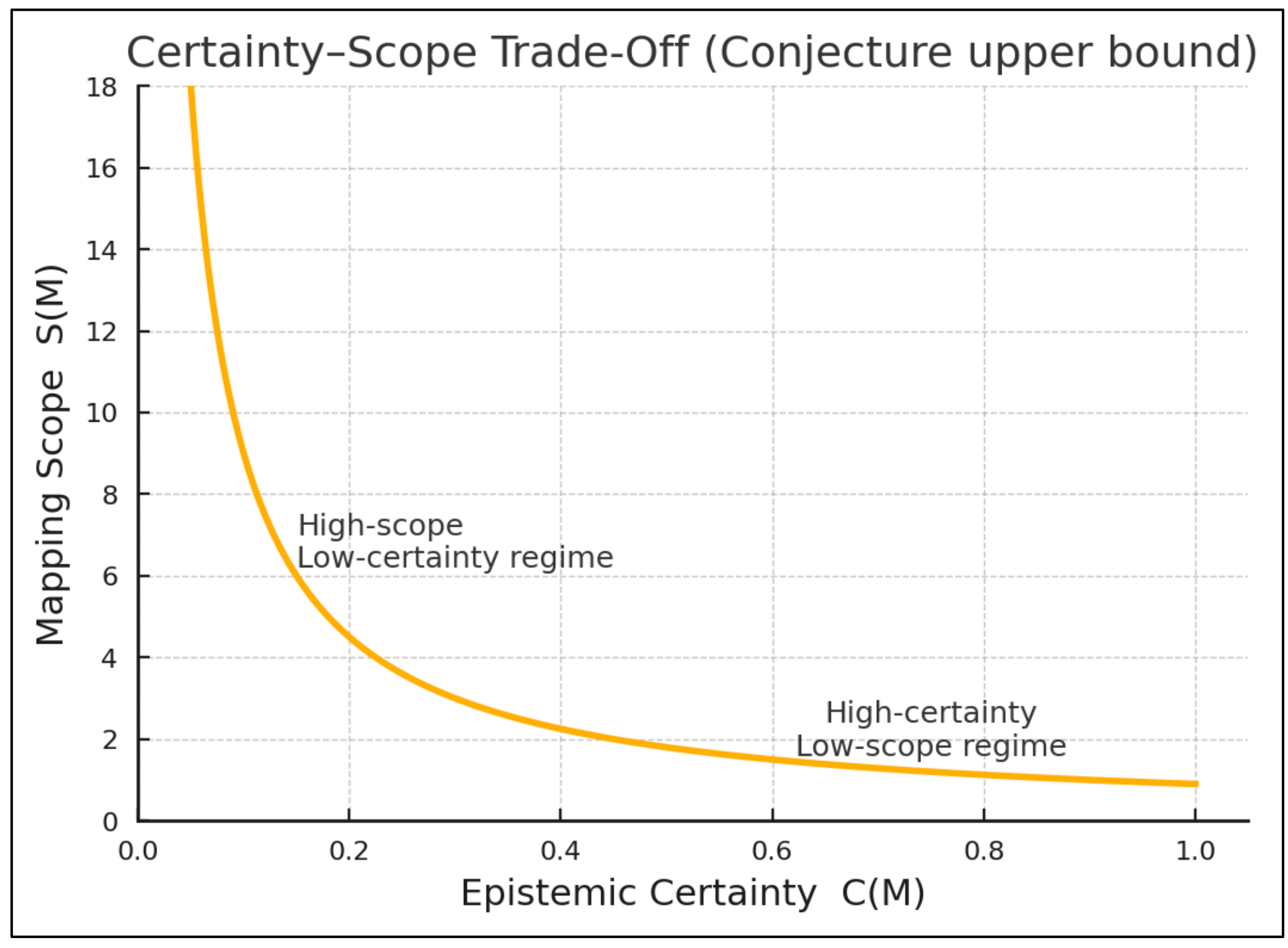

*Figure 2. The hypothetical shape of the Certainty–Scope trade-off curve represents the constraint* $C(M) \cdot S(M) \leq k$. *The x-axis represents epistemic certainty C(M), and the y-axis represents mapping scope S(M). For real systems, the operational point lies somewhere below the curve, depending on task constraints and architecture.*

Two immediate corollaries follow:

a) Provable-certainty regime: if $C(M) = 1$, then $S(M) \leq k$. Thus, achieving perfect certainty necessarily restricts the scope of the AI system to at most $k$.
b) Wide-scope regime: conversely, if $S(M) > k$ then $C(M) < 1$. Thus, no sufficiently expressive system can achieve absolute certainty over a very rich domain. Any system whose operational breadth exceeds the threshold $\boldsymbol{k}$ incurs an irreducible error (or hallucination) rate that no amount of training or engineering can eliminate.

This conjecture is currently only a theoretical inequality. It may ultimately be derivable from first principles, for example, via information theory or computational complexity arguments, or it may be an empirical regularity or contingent truth observable in practice but not formally universal. Either way, it provides a succinct formulation of the trade-off that AI practitioners have often intuited, as we shall see in the next section.

## 3. Grounding the conjecture

The conjecture introduced in this article, while original in its explicit formalisation, does not seem very controversial, as it builds upon an extensive landscape of research that implicitly supports its core intuition. Some notable strands of literature converge to ground and contextualise it. I summarise them here briefly because they are well-known.

*Expressiveness versus Tractability.* In knowledge representation and symbolic AI, a long-recognised trade-off exists between the *expressiveness* of a logical language and the computational *tractability* of reasoning within that language (Brachman and Levesque 1987). Highly expressive formalisms enable richer and broader representational capacity but typically lead to computationally intractable reasoning tasks, often resulting in undecidability or exponential complexity. Conversely, simpler formalisms allow efficient, even guaranteed, reasoning at the expense of representational depth and domain coverage. This fundamental tension between the breadth of expressiveness and the strength of inferential guarantees aligns closely with the conjecture's proposal that broad data-mapping capacity and flawless correctness of outputs are fundamentally in opposition.

*No Free Lunch Theorems.* Complementing insights from logic-based AI, the No Free Lunch (NFL) theorems in machine learning (ML) provide robust theoretical backing to the conjecture's claim about unavoidable trade-offs. Wolpert and Macready (1997) demonstrated that no learning algorithm universally outperforms others across all conceivable tasks, thus indicating that absolute generality inevitably sacrifices guaranteed performance on specific tasks unless specialised assumptions or constraints are embedded. The NFL theorems underscore the impossibility of achieving both broad task-generalisation and maximal performance simultaneously, mirroring the conjecture assertion that increasing scope inevitably diminishes epistemic certainty.

*Limits of Formal Verification.* The literature on formal methods and AI safety highlights the severe limitations of formal verification for complex, open-world systems. Classical results, such as the Halting Problem and Rice's theorem, underscore fundamental limitations in verifying arbitrary computational systems. This limitation is echoed in recent work by Dickson (2024), who surveys the infeasibility of formal verification for modern, opaque AI models, arguing that safety guarantees are often only achievable in simplified simulation environments. Similarly, Seshia, Sadigh, and Sastry (2022) propose layered assurance architectures precisely because complete formal verification is infeasible for systems interacting with dynamic, unstructured environments. These insights support the core claim of the conjecture: formal correctness scales only with strict constraints on environmental complexity.

*Probably Approximately Correct* (PAC) learning theory. Further support for the conjecture arises from PAC, which quantifies trade-offs between generalisability and the size of hypothesis classes (Valiant 1984). In PAC frameworks, achieving low error with high confidence requires several samples that grow with the complexity of the target concept class, often measured by VC dimension or other capacity metrics. This reflects a broader principle: more expressive or wide-ranging learners require more information to guarantee correctness, reinforcing the conjecture that broader operational domains entail higher epistemic risk.

Collectively, these strands of research from symbolic AI, ML theory, and formal verification ground the conjecture within established theoretical frameworks and empirical observations, indicating that it articulates a fundamental, previously implicit trade-off explicitly and rigorously. As such, the conjecture synthesises and elevates established insights, framing them clearly for future philosophical, theoretical, and practical exploration.

## 4. Some logical implications

The conjecture, if correct, has several implications. Here, I analyse five of the most interesting.

*4.1. Limits on verification strategies*

The conjecture clarifies why deductive verification fails to scale beyond comparatively simple domains. As tasks or environments grow in complexity—that is, as $\boldsymbol{S}(\boldsymbol{M})$ increases—purely logical proof methods inevitably falter. This accords with long-standing results: verifying arbitrary software or AI programmes in full generality is undecidable, and practical verification succeeds only by narrowing the problem space or adopting simplifying assumptions (Pearl 1988). Hence, complete formal guarantees remain achievable solely for AI systems confined to tightly bounded worlds, necessarily limiting their functional scope and power.

*4.2. Architectural bifurcation*

If the conjecture is correct, it suggests a principled division of labour in system design. A well-established direction is the hybrid architecture illustrated in

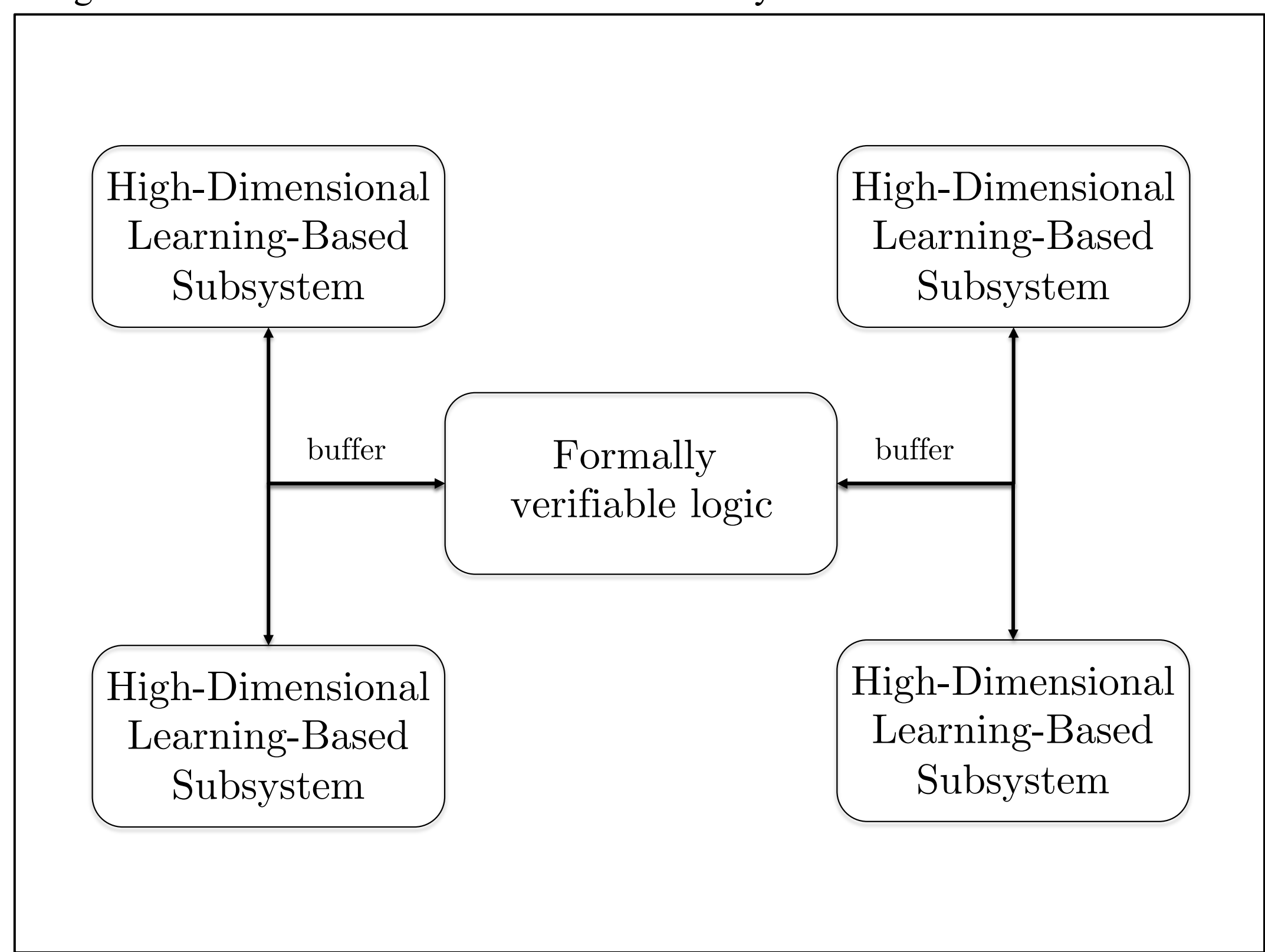

Figure 3: a composition of high-$\boldsymbol{C}$, low-$\boldsymbol{S}$ kernels (e.g., formally verified controllers and other safety-critical cores) surrounded by low-$\boldsymbol{C}$, high-$\boldsymbol{S}$ envelopes (e.g., perceptual front-ends, large generative models, or planners that handle unstructured inputs). By acknowledging that no single module can possess both a broad scope and deductive certainty, the ensemble balances these two through carefully designed interfaces. Emerging agentic AI frameworks may be converging on this layered pattern.

For instance, an autonomous vehicle might combine a high-assurance braking controller, verified for safety under defined parameters, with a deep neural network-based perception system for interpreting camera data. The verified component guarantees correctness under known conditions, while the perception system tolerates some errors to handle real-world variability.

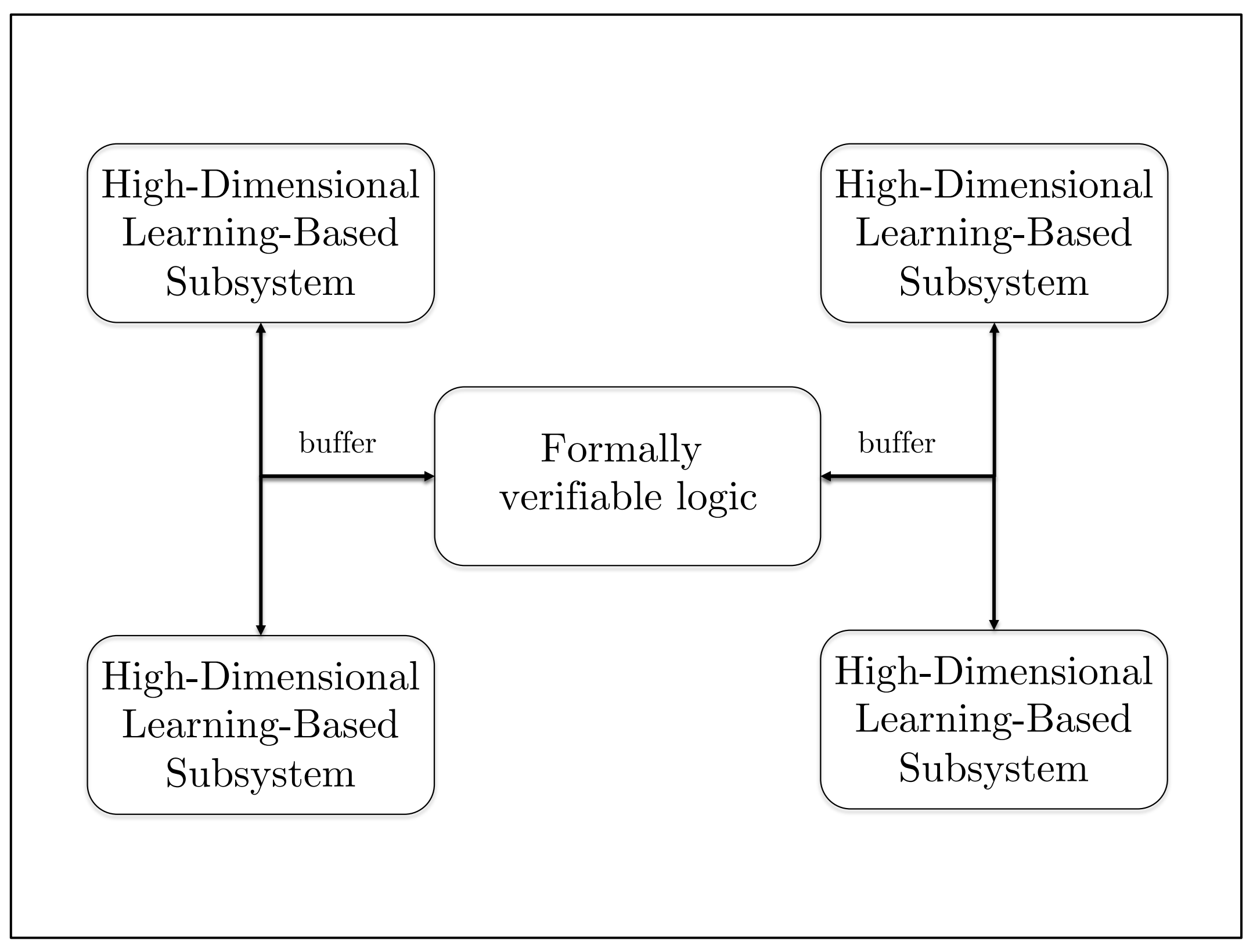

*Figure 3 Illustration of a hybrid structure: a central core (formally verifiable logic) is surrounded by high-dimensional, learning-based subsystems. Arrows represent data flow, with error-absorbing buffers or fallback mechanisms between probabilistic and deterministic zones.*

This theoretical dichotomy is mirrored in empirical systems. For example, symbolic reasoning tools such as SAT solvers, type checkers, or certified kernel modules can offer full correctness guarantees, but only within precisely defined formal languages or execution environments (low $\boldsymbol{S}(\boldsymbol{M})$, high $\boldsymbol{C}(\boldsymbol{M})$). By contrast, large language models, such as the GPT series, generate fluent and flexible responses across many domains (high $\boldsymbol{S}(\boldsymbol{M})$), but unavoidably commit occasional factual or logical errors, and their correctness cannot be proven a priori (low $\boldsymbol{C}(\boldsymbol{M})$). These examples concretely instantiate the trade-off space outlined by the conjecture.

A further illustration of this trade-off appears in the contrast between Coq and AlphaCode. Coq is a formal proof assistant that requires users to construct explicitly machine-checkable correctness proofs. Programs verified in Coq are provably correct relative to their specifications. Still, they can only be developed within a strictly formalised language and logic framework, an example of low $\boldsymbol{S}(\boldsymbol{M})$ and high $\boldsymbol{C}(\boldsymbol{M})$ (Bertot and Castéran 2004). AlphaCode, developed by DeepMind, uses transformer-based architectures to generate solutions to programming problems across diverse domains. It shows impressive scope, tackling novel problems in multiple languages without task-specific retraining. However, its outputs require validation through unit tests or human review, and there is no formal guarantee of correctness, a hallmark of high $\boldsymbol{S}(\boldsymbol{M})$ and low $\boldsymbol{C}(\boldsymbol{M})$ (Li et al. 2022). These systems exemplify the certainty–scope trade-off in operational practice: Coq offers ironclad assurance within tightly constrained domains, while AlphaCode excels in breadth but offers no deductive guarantees.

### *4.3. Evaluation metrics*

The conjecture also cautions against misinterpretation of performance metrics. If a benchmark or competition claims that a system achieves both broad, human-level competence across many tasks and near-perfect accuracy, one should suspect a violation of the certainty–scope trade-off. Either the task scope is effectively narrower than it appears, or the reported accuracy is not backed by rigorous proof and might not hold in worst-case scenarios. This insight encourages the design of new metrics that explicitly account for the scope of inputs alongside accuracy. For example, evaluators might measure performance as a pair *(coverage, confidence)* rather than a single score, reflecting the fact that pushing for greater coverage of situations will inevitably introduce some areas of uncertainty or errors.

### *4.4. Regulatory and ethical design*

In safety-critical AI applications, regulators often desire complete guarantees of correctness, e.g., “no accidents” from an AI-based, autonomous vehicle. The conjecture suggests that such demands are *in tension* with requiring the AI to handle the full open-world complexity of real roads and traffic. Policies predicated on 100% correctness in an open domain implicitly assume the conjecture can be circumvented or disproved. However, in either case, the burden of proof lies with those who propose the policies. Given the plausible nature of the conjecture, a more realistic regulatory approach would be to require quantifiable and manageable risk, acknowledging an error rate but governing it through standards of prudent risk management (Floridi 2021). Ethically, this viewpoint urges honesty about the remaining uncertainty in any advanced AI system. Rather than promising perfect safety, developers would implement oversight mechanisms, e.g., fallbacks, insurance, and periodic audits, to mitigate the *irreducible* errors that come with broad functionality.

### *4.5. Research horizons*

Proving or disproving the conjecture presents an exciting challenge that spans several fields. Techniques from information theory (e.g., limits on compression and generalisation), computational learning theory (probably approximately correct learning bounds, no-free-lunch theorems), and proof complexity could be brought together to either derive the inequality from deeper axioms or to identify the conditions under which it fails. Even attempts to formalise the principle as one or more theorems to be proved could yield new insights. For example, one might discover tighter bounds or alternative formulations; perhaps there is a different function relating error and scope more directly. Falsifying the conjecture would be even more interesting. In particular, a constructive counter-example provided by an AI system that achieves both essentially perfect reliability *and* wide-ranging competence, would point toward a fundamentally new kind of architecture (or a new understanding of "scope" or "certainty"). Either outcome—proof or counterproof—would be profoundly informative for the future of AI design.

### *5. Some philosophical consequences*

The proposed conjecture echoes a long philosophical line of research connecting knowledge, risk, and the limits of formal representation. In epistemology, the doctrine of underdetermination holds that empirical data under-specify the theories or models that can explain them (Quine 1969). Similarly, in our context, rich, high-entropy inputs underdetermine any complete deductive guarantee a system might offer, as there are always many possible interpretations or outcomes, so an AI processing unrestricted reality cannot pin down a single assured correct response in every case. Conversely, a logically perfect proof operates only within what may be called, in Wittgensteinian terms, a closed language-game, insulated from the "semantic turbulence" of everyday meaning and experience (Wittgenstein (1953). The conjecture instantiates this idea in engineering terms: it is a structural analogue of Popper's dilemma in the philosophy of science, whereby universal statements about the empirical world can never be finally verified but only provisionally corroborated within error bounds (Popper 1959). Here, an AI's universal claim to never err is perennially tentative, tested only to a specific level of confidence.

Ethically, the conjecture highlights the importance of prudence and honesty in addressing uncertainty. If the conjecture is correct, design teams that market open-world AI systems while implying total reliability are epistemically overreaching, to say the least; they risk creating a moral hazard, as users may assume no responsibility and no possibility of error. In AI deployment, designers and users alike should consider that a complex system is intrinsically fallible and plan accordingly. Acknowledging this does not entail paralysis or lack of ambition. Instead, it encourages layered assurance strategies: critical tasks can be handled by a reliable core—perhaps a formally verified algorithm for a narrowly defined function—while more complex and uncertain decisions are made by probabilistic modules whose outputs are monitored and subject to checks. Such hybrid designs accept a degree of uncertainty yet manage it in a principled way, aligning with an ethics of responsibility rather than a false promise of infallibility.

The conjecture also reframes debates on explainability and accountability in AI. If truly broad-scope systems are unavoidably fallible, then the quest for a single definitive explanation—which goes beyond a general assessment—for each of their errors or outputs may be misguided. In complex systems, no single factor fully determines an outcome; there will always be a statistical or probabilistic element. What matters instead is providing sufficiently informative and truthful rationales that enable stakeholders to understand the system's operation in general terms and to evaluate residual risks. In this sense, the conjecture aligns technical reality with an ethics of transparency. Rather than aiming for total epistemic certainty, which may be unattainable in principle, one should aim for clear communication of uncertainty and assumptions. Stakeholders, from engineers and regulators to end-users, can then make informed and more prudent decisions, knowing both the capabilities and the inevitable limitations of AI systems.

## 6. Conclusion

The conjecture formalises a fundamental trade-off at the heart of AI design: one cannot maximise an AI's scope (its ability to handle the open-ended complexity of the real world) and simultaneously maximise its certainty (absolute, provable correctness). While this tension is widely recognised informally, the conjecture introduced here provides, for the first time, a formalised, testable hypothesis that explicitly unifies existing theoretical and empirical insights into a coherent principle. By making this previously implicit trade-off explicit and open to rigorous verification, the conjecture significantly reframes both engineering ambitions and philosophical expectations for AI. It highlights that demands for perfect accuracy must be tempered by an understanding of an AI's domain of operation, and, conversely, that expanding an AI's domain inevitably introduces some uncertainty.

The conjecture is not an isolated speculation; it is linked to established theoretical insights, ranging from the trade-offs between expressiveness and tractability in logic, to the No Free Lunch limits on universal learning algorithms, to practical observations that complex AI systems resist complete formal verification. This is also reinforced by PAC learning bounds, which quantify how increasing the complexity of the hypothesis space drives the sample demands for approximate correctness. Such parallels suggest that the conjecture captures a fundamental, deep constraint, one that is already hinted at by different but converging research threads.

Looking ahead, whether future work proves, refines, or overturns this conjecture, its significance will be profound. A formal proof would establish a new law-like constraint on AI capabilities, guiding researchers toward hybrid architectures and cautioning against epistemically over-ambitious claims. A refinement might indicate how much certainty must be traded off for a given scope (and vice versa), providing quantitative guidance to system designers. An outright refutation—including an example of an AI that breaks the presumed inequality—would be the most startling outcome, revealing that our current paradigms have not exhausted what is possible. Any of these outcomes would mark a leap in our understanding. In the meantime, the explicit articulation of the conjecture may serve a valuable purpose: it gives AI researchers, theorists, and policymakers a clear target for debate and investigation. Rather than intuiting this trade-off in informal ways, we can now analyse it directly and refine it. By confronting the challenge implicit in this conjecture, we move toward AI systems that are not just more powerful, but also more principled in how they balance the timeless dialectics between breadth of capability and reliability of performance.

## Acknowledgements

I am very thankful to Claudio Novelli, for pointing out that the original Figure 2 was incorrect (very obviously, once he drew my attention to it); to Alberto Messina, for showing me that the formulation of the conjecture published in a previous version of the article, contained mistakes; and to David Watson, for highlighting some linguistic imprecisions. One may wonder whether I got anything right the first time. On a more constructive note, they are the best reminder of the value of collaboration.

## References

Bertot, Yves, and Pierre Castéran. 2004. Interactive Theorem Proving and Program Development: Coq'Art: The Calculus of Inductive Constructions. Berlin: Springer.

Brachman, Ronald J., and Hector J. Levesque. 1987. "Expressiveness and Tractability in Knowledge Representation." *Computational Intelligence* 3 (4): 78–93.

Bubeck, Sébastien, *et al.* 2023. "Sparks of Artificial General Intelligence: Early Experiments with GPT-4." arXiv preprint arXiv:2303.12712.

Chaitin, Gregory J. 1969. "On the Length of Programs for Computing Binary Sequences." *Journal of the ACM* 13 (4): 547–569.

Dickson, Andrew. 2024. "Limitations on Formal Verification for AI Safety." AI Alignment Forum, 19 Aug 2024. *(Accessed online).*

Floridi, Luciano. 2021. *The Logic of Information.* Oxford: Oxford University Press.

Goodfellow, Ian, Yoshua Bengio, and Aaron Courville. 2016. *Deep Learning.* Cambridge, MA: MIT Press.

Kolmogorov, Andrei N. 1965. "Three Approaches to the Quantitative Definition of Information." *Problems of Information Transmission* 1 (1): 1–7.

Li, Yujia, David Choi, Jarin Timilsina, Laurent Sifre, et al. 2022. "Competition-Level Code Generation with AlphaCode." arXiv preprint arXiv:2203.07814. https://arxiv.org/abs/2203.07814

Pearl, Judea. 1988. *Probabilistic Reasoning in Intelligent Systems.* San Mateo, CA: Morgan Kaufmann.

Popper, Karl. 1959. *The Logic of Scientific Discovery.* London: Hutchinson.

Quine, W. V. O. 1969. "Epistemology Naturalised." In *Ontological Relativity and Other Essays*, 69–90. New York: Columbia University Press.

Russell, Stuart, and Peter Norvig. 2022. *Artificial Intelligence: A Modern Approach (4th ed.).* Harlow, UK: Pearson.

Seshia, Sanjit A., Dorsa Sadigh, and S. Shankar Sastry. 2022. "Toward Verified Artificial Intelligence." *Communications of the ACM* 65 (7): 36–38.

Shannon, Claude E. 1948. "A Mathematical Theory of Communication." *Bell System Technical Journal* 27 (3): 379–423.

Wittgenstein, Ludwig. 1953. *Philosophical Investigations.* Oxford: Blackwell.

Wolpert, David H., and William G. Macready. 1997. "No Free Lunch Theorems for Optimization." *IEEE Transactions on Evolutionary Computation* 1 (1): 67–82.